\documentclass{article}
\usepackage{spconf,amsmath,graphicx,amssymb}
\usepackage[export]{adjustbox}
\usepackage{subfig}
\usepackage{color}
\usepackage{enumitem}
\usepackage{multirow}
\usepackage{bm,upgreek}
\PassOptionsToPackage{hyphens}{url}\usepackage{hyperref}

\hypersetup{colorlinks=true}
\DeclareMathOperator*{\argmin}{argmin}
\DeclareMathOperator*{\argmax}{argmax}

\title{Links: A High-Dimensional Online Clustering Method}

\name{Philip Andrew Mansfield\textsuperscript{1} \quad Quan Wang\textsuperscript{1} \quad Carlton Downey\textsuperscript{2} \quad Li Wan\textsuperscript{1} \quad Ignacio Lopez Moreno\textsuperscript{1}}

\address{\textsuperscript{1}Google Inc., USA \qquad \textsuperscript{2}Carnegie Mellon University, USA\\[4pt] {
  \normalsize
  \textsuperscript{1}
    \{
    \href{mailto:memes@google.com}{\nolinkurl{memes}},
    \href{mailto:quanw@google.com}{\nolinkurl{quanw}},
    \href{mailto:liwan@google.com}{\nolinkurl{liwan}},
    \href{mailto:elnota@google.com}{\nolinkurl{elnota}}
    \}
  {\tt @google.com} \qquad
  \textsuperscript{2}
  \href{mailto:cmdowney@cs.cmu.edu}{\nolinkurl{cmdowney@cs.cmu.edu}}
}}

\begin{document}
\ninept
\maketitle
\begin{abstract}
We present a novel algorithm, called \textit{Links}, designed to perform online clustering on unit vectors in a high-dimensional Euclidean space. The algorithm is appropriate when it is necessary to cluster data efficiently as it streams in, and is to be contrasted with traditional batch clustering algorithms that have access to all data at once. For example, Links has been successfully applied to embedding vectors generated from face images or voice recordings for the purpose of recognizing people, thereby providing real-time identification during video or audio capture.
\end{abstract}
\section{Introduction}
\label{sec:intro}

Although a wide selection of clustering methods are available \cite{everitt2011cluster,hennig2015handbook}, most of them assume concurrent access to all data being clustered.  Our interest is in efficiently clustering each datum as it becomes available, for applications that require unsupervised learning in real time.

The Links approach is to estimate the probability distribution of each cluster based on its current constituent vectors, to use those estimates to assign new vectors to clusters, and to update estimated distributions with each added vector. The update step includes fixing past cluster assignments where indicated by taking the additional data into account, although this is primarily to improve the internal model over time, since in typical online usage scenarios, each cluster assignment is provided once, at the time a new vector is made available.

Prior work \cite{straub2015small} addressing online clustering of unit vectors employs a small-variance approximation and is applied to low-dimensional problems such as segmentation of surface normals in 3D. Our approach is complementary in that it uses a high-dimensional approximation, and has been applied to problems with relatively high variance.

Links has been used to cluster CNN-based FaceNet embeddings \cite{schroff2015facenet} and LSTM-based voice embeddings \cite{wan2017ge2e}. The results of the latter experiment are presented in a separate paper \cite{wang2017speaker}. The current paper focuses on the technical details of the algorithm.

\section{Model}
\label{sec:model}

\subsection{Generative model for a cluster}
\label{sec:generative}

Let $X=\{\mathbf{x}_i\}$ be a set of unit-length vectors in $\mathbb{R}^{N}$. They are confined to the submanifold $S^{N-1}$, and to determine proximity for the purpose of clustering these vectors, we will use the natural metric on this submanifold, which is simply the angle between vectors:
\begin{equation}
\angle(\mathbf{x},\mathbf{x'}) = \arccos(\mathbf{x}\cdot\mathbf{x'}).
\end{equation}

We address the problem of cluster distributions within this submanifold with the following properties:
\begin{enumerate}
  \item Each cluster has a \textit{center} vector $\bm{\upmu}$ and its member vectors $\mathbf{x}$ are generated by a probability density that is isotropic in the sense that it only depends on distance from the center, $\mathbf{x}\sim\rho(\mathbf{x};\bm{\upmu})=f\left(\angle(\mathbf{x},\bm{\upmu})\right)$.
  \item The  function $f$ is the same for every cluster, so that probability densities for different clusters are related by isometry.
  \item $f(\theta)$ decreases exponentially with $\theta$; for example, as a Gaussian suitably normalized on $S^{N-1}$:
  \begin{equation}
  f(\theta)\propto \mbox{\Large$e^{-\frac{\theta^2}{2\sigma^2}}$}.
  \end{equation}
  This ensures that the distribution is reasonably localized, since the exponential decrease compensates for a polynomial factor in the marginal distribution of $\theta$:
  \begin{equation}
  \rho(\angle(\mathbf{x},\bm{\upmu})=\theta)=\mathcal{A}(\sin\theta)^{N-2}f(\theta)
  \end{equation}
  where $\mathcal{A}$ is a constant equal to the hypersurface area of $S^{N-2}$,
  \begin{equation}
  \mathcal{A}=\frac{2\pi^{(N-1)/2}}{\Gamma\left(\frac{N-1}{2}\right)}.
  \end{equation}
  \item The prior distribution $\rho(\bm{\upmu})$ for the center of a cluster $\bm{\upmu}$ is constant on $S^{N-1}$ (no unit vector is preferred).
\end{enumerate}

\subsection{Estimated distribution}
\label{sec:estimated}

Given a set $X=\{\mathbf{x}_i\}$ chosen randomly from the same cluster, but without knowledge of the center of the cluster, we would like to estimate the cluster's probability distribution. The likelihood of the center value $\bm{\upmu}$ is
\begin{equation}\label{eqn:likelihood}
\begin{split}
L(\bm{\upmu};X) & =\prod_i \rho(\mathbf{x}_i;\bm{\upmu}) \\
& \propto \mbox{\Large$e^{-\frac{\sum_i \left(\angle(\mathbf{x}_i,\bm{\upmu})\right)^2}{2\sigma^2}}$}
\end{split}
\end{equation}
Since the prior $\rho(\bm{\upmu})$ is constant, the posterior $\rho(\bm{\upmu}|X)$ is also proportional to the expression in equation \ref{eqn:likelihood}. The maximum likelihood (and maximum a posteriori) center is therefore
\begin{equation}
\bm{\hat{\upmu}}=\argmin_{\bm{\upmu}}\sum_i \big(\angle(\mathbf{x}_i,\bm{\upmu})\big)^2
\end{equation}
which is the same as the centroid of the vectors $\{\mathbf{x}_i\}$ as defined for a hypersphere according to \cite{buss2001spherical}. The estimated probability distribution for the cluster is
\begin{equation}
\begin{split}
\hat{\rho}(\mathbf{x};X) & \propto \int_{S^{N-1}}  L(\bm{\upmu};X)\rho(\mathbf{x};\bm{\upmu})d\bm{\upmu} \\
& \propto \int_{S^{N-1}}  \mbox{\Large$e^{-\frac{\left(\angle(\mathbf{x},\bm{\upmu})\right)^2+\sum_i \left(\angle(\mathbf{x}_i,\bm{\upmu})\right)^2}{2\sigma^2}}$} d\bm{\upmu}.
\end{split}
\end{equation}
The probability that a new vector $\mathbf{x}$ belongs to the same cluster can then be estimated as the cumulative amount
\begin{equation}\label{eqn:cumulative}
\int_{\{\mathbf{y}\in S^{N-1}\,\mid\,\hat{\rho}(\mathbf{y};X)\geq\hat{\rho}(\mathbf{x};X)\}} \hat{\rho}(\mathbf{y};X)d\mathbf{y}.
\end{equation}

\subsection{High-dimensional approximation}
\label{sec:approximation}

Our primary interest is in problems with relatively large $N$. For example, our typical embedding vectors have $N\ge128$. For large enough $N$, the following are true:

\begin{description}
  \item [Lemma 1] Two randomly chosen vectors $\mathbf{x},\mathbf{x'}$ are almost always almost perpendicular, i.e.,
  \begin{equation}
  P(\mathbf{x}\cdot\mathbf{x'}>\delta)<\epsilon
  \end{equation}
  for some positive numbers $\delta\ll1$ and $\epsilon\ll1$.
  \item [Lemma 2] The angle $\theta$ between a cluster center and a random vector from that cluster is almost always almost equal to a global constant $\theta_c$, i.e.,
  \begin{equation}\label{key}
  P\left(|\theta-\theta_c|>\delta\right)<\epsilon
  \end{equation}
  for some positive numbers $\delta\ll\pi$ and $\epsilon\ll1$.
  \item [Lemma 3] Given two randomly chosen vectors from a cluster with center $\bm{\upmu}$, their components perpendicular to $\bm{\upmu}$ will almost always be almost perpendicular to each other, i.e.,
  \begin{equation}
  P\Big(\big(\mathbf{x}-(\mathbf{x}\cdot\bm{\upmu})\bm{\upmu}\big)\cdot\big(\mathbf{x'}-(\mathbf{x'}\cdot\bm{\upmu})\bm{\upmu}\big)>\delta\Big)<\epsilon
  \end{equation}
  for some positive numbers $\delta\ll1$ and $\epsilon\ll1$.
\end{description}

To assess whether to add a new vector $\mathbf{x}$ to an existing cluster known to include the $k$ vectors $\left\lbrace\mathbf{x}_i\right\rbrace_{i=1}^k$, we determine a threshold $\mathbf{x}\cdot\bm{\hat{\upmu}}\geq s(k)$ on the cosine similarity between the new vector and the centroid $\bm{\hat{\upmu}}$ of the existing vectors. Using the approximation in lemmas 2 and 3, and assuming $N\gg k$, we can compute vector components in an orthonormal basis including $\bm{\upmu}$, $\frac{1}{\sin\theta_c}(\mathbf{x}-\cos\theta_c\bm{\upmu})$ and $\left\lbrace \frac{1}{\sin\theta_c}\left(\mathbf{x}_i-\cos\theta_c\bm{\upmu}\right)\right\rbrace_{i=1}^k$. This yields

\begin{equation}
\bm{\hat{\upmu}}=\frac{1}{\sqrt{k^2\cos^2\theta_c+k\sin^2\theta_c}}\sum_{i=1}^{k}\mathbf{x}_i
\end{equation}
and a threshold of
\begin{equation}\label{eqn:single_threshold}
\begin{split}
s(k) & =\frac{k\cos^2\theta_c}{\sqrt{k^2\cos^2\theta_c+k\sin^2\theta_c}} \\
& =\frac{T_c^2}{\sqrt{\frac{1}{k}+\left(1-\frac{1}{k}\right)T_c^2}}
\end{split}
\end{equation}
where $T_c=\cos\theta_c$, which we call the \textit{cluster similarity threshold}.

Note that
\begin{equation}
\lim_{k\rightarrow\infty}\bm{\hat{\upmu}}=\bm{\upmu}
\end{equation}
and
\begin{equation}
\lim_{k\rightarrow\infty}s(k)=T_c,
\end{equation}
which confirms that as we accumulate more vectors in a given cluster, the center and cosine similarity threshold of the \textit{estimated} distribution approach the center and cosine similarity threshold of the \textit{generative} distribution (i.e., the estimate improves). Since $s(k)$ is a strictly increasing function of $k$, the variance of the estimated distribution decreases with $k$.

Similarly, to assess whether two clusters are the same, we determine a threshold on the cosine similarity between their centroids $\bm{\upmu}_c\cdot\bm{\upmu}_c'\geq s(k,k')$ where, for $N\gg k$ and $N\gg k'$,
\begin{equation}\label{eqn:multi_threshold}
\begin{split}
s(k,k') & =\frac{1}{\sqrt{\left( 1+\frac{1}{k}\tan^2\theta_c\right) \left( 1+\frac{1}{k'}\tan^2\theta_c\right)}} \\
& =\frac{1}{\sqrt{\bigg( 1+\frac{1}{k}\left(\frac{1}{T_c^2}-1\right)\bigg)\bigg(1+\frac{1}{k'}\left(\frac{1}{T_c^2}-1\right)\bigg)}}.
\end{split}
\end{equation}

Note that equation \ref{eqn:single_threshold} is the special case with $k'=1$,
\begin{equation}
s(k,1)=s(k),
\end{equation}
and
\begin{equation}\label{eqn:multi_limit}
\lim_{k,k'\rightarrow\infty}s(k,k')=1.
\end{equation}
The latter confirms that the centers estimated from the two sets of cluster points converge.

\section{Algorithm}
\label{sec:algorithm}

\subsection{Online clustering}
\label{sec:online}

Each new input vector is assigned to a cluster as soon as it is produced, with no knowledge of future vectors and no backtracking. A unique ID for that cluster is returned. The clusterer keeps statistical information about the vectors received so far.  Although it cannot change a previous answer, it can change the internal representation of cluster statistics, such as improvements to estimated distributions as well as cluster splits and merges when indicated by new information.

\subsection{Internal representation}
\label{sec:internal}

The Links algorithm's internal representation is a two-level hierarchy: clusters are collections of subclusters, and subclusters are collections of input vectors.  The subclusters are represented as nodes in a graph whose edges join `nearby' nodes (meaning subclusters that likely belong to the same cluster given the data so far), and clusters are defined as connected components of the graph.  Whereas subclusters are indivisible, clusters can become split along graph edges in response to changes in subcluster estimated probability distributions as new data is added. Alternatively, subclusters joined by an edge can become merged in response to changes.

The reasons for maintaining this two-level hierarchy (rather than, say, an arbitrary number of levels) are efficiency and practicality. It is efficient because the algorithm scales with number of subclusters rather than number of vectors.  It is practical because the key cluster substructure that can affect future cluster IDs is the set of potential split points.

\subsection{Assessing cluster membership}
\label{sec:assessing}

When a new vector $\mathbf{x}$ is available, compute its cosine similarity to each subcluster centroid $\bm{\hat{\upmu}}_j$, and add it to the most-similar subcluster if the similarity is above a fixed threshold $T_s$.
In other words, let
\begin{equation}
J = \argmax_j\{\mathbf{x}\cdot\bm{\hat{\upmu}}_j\}.
\end{equation}
If
\begin{equation}\label{eqn:subcluster}
\mathbf{x}\cdot\bm{\hat{\upmu}}_J\geq T_s
\end{equation}
then add $\mathbf{x}$ to subcluster $J$. $T_s$, called the \textit{subcluster similarity threshold}, is a hyperparameter determining the granularity of cluster substructure appropriate for the data.

If inequality \ref{eqn:subcluster} does not hold, then start a new subcluster containing just $\mathbf{x}$. Next, use the estimated probability distribution of subcluster $J$ to determine whether to include the new subcluster in the same cluster as $J$, by thresholding the cumulative probability in expression \ref{eqn:cumulative}. In the high-dimensional approximation, this means the subcluster is included in the cluster whenever
\begin{equation}\label{eqn:cluster}
\mathbf{x}\cdot\bm{\hat{\upmu}}_J\geq s(k_J)
\end{equation}
where $k_J$ is the number of vectors in the subcluster $J$. To a first approximation, $s(k)$ is as given in equation \ref{eqn:single_threshold}. This will be further refined in section \ref{sec:anisotropy}. If inequality \ref{eqn:cluster} does hold, then add an edge to the graph joining the new subcluster to subcluster $J$.

\subsection{Updating clusters}
\label{sec:updating}

When a new vector is added to an existing subcluster, the subcluster's centroid may change. If this brings it within the subcluster similarity threshold of the centroid of another subcluster currently joined to the first by an edge, then the two are merged. In other words, if $\bm{\hat{\upmu}}_i\cdot\bm{\hat{\upmu}}_j\geq T_s$, then nodes $i$ and $j$ are replaced with a single node containing the vectors of both, and with the edge connections of both. Since the merging process also results in a new subcluster centroid, this check is continued recursively on affected subclusters.

Next, the edges joining affected nodes are checked for validity. The edge joining subclusters $i$ and $j$ is removed if the following does not continue to hold:
\begin{equation}\label{eqn:no_split}
\bm{\hat{\upmu}}_i\cdot\bm{\hat{\upmu}}_j\geq s(k_i,k_j)
\end{equation}
where $s(k_i,k_j)$ is approximately as given in equation \ref{eqn:multi_threshold}, but with improvements to follow in section \ref{sec:anisotropy}. After severing a cluster in two by removing an edge, an attempt is made to re-join the two parts by adding an edge from the affected node to a new partner node that does satisfy inequality \ref{eqn:no_split}. If no such partner is found, then the cluster remains permanently split.

\subsection{Anisotropy}
\label{sec:anisotropy}

Equations \ref{eqn:single_threshold} and \ref{eqn:multi_threshold} were used to determine thresholds for membership in the same cluster as a given subcluster, effectively treating the subcluster's members as randomly chosen from the cluster and not correlated with each other. If one were to properly take into account intra-subcluster correlations, then one consequence is that the limit in equation \ref{eqn:multi_limit} would be reduced to a positive number $T_p<1$, which we call the \textit{pair similarity maximum},
\begin{equation}\label{eqn:anisotropic_limit}
\lim_{k,k'\rightarrow\infty}s(k,k')=T_p,
\end{equation}
whereas the value of $s(1,1)$, which is $T_c^2$, would remain unchanged. Any implicit anisotropy in the cluster distribution, such as an elongation along a preferred axis, will further reduce the value of $T_p$ without changing $s(1,1)$. A simple though approximate way to incorporate these adjustments into the algorithm is to replace $s(k,k')$ and $s(k)$ by the following interpolated versions:
\begin{equation}\label{eqn:anisotropic_multi}
\tilde{s}(k,k')=T_c^2+\frac{T_p-T_c^2}{1-T_c^2}\left(s(k,k')-T_c^2\right)
\end{equation}
\begin{equation}\label{eqn:anisotropic_single}
\tilde{s}(k)=\tilde{s}(k,1).
\end{equation}

\subsection{Hyperparameter Tuning}
\label{sec:hyperparameter}

The similarity thresholds $T_c$, $T_s$ and $T_p$ need to be tuned to best represent the data source. This is done by manually labeling a dataset with cluster IDs, running the clusterer on the data, and adjusting hyperparameters to improve the accuracy of the output cluster IDs. Accuracy is simply fraction of correct IDs. Prior to evaluation, the Hungarian algorithm \cite{kuhn1955hungarian} is used to map a subset of output cluster IDs bijectively to a subset of ground truth cluster IDs in such a way that produces the best possible accuracy. For some applications an alternate objective has been used; for example, one that gives different weights for conflating IDs vs. fracturing IDs, to reflect the seriousness of each type of error in practise.

\section{Acknowledgements}
\label{sec:ack}

The authors would like to thank Dr. Brian Budge and Dr. Navid Shiee for help with APIs and evaluation frameworks used in the implementation of the Links algorithm.

\newpage
\bibliographystyle{IEEEbib}
\bibliography{supplement_refs}

\begin{thebibliography}{1}

\bibitem{everitt2011cluster}
Brian Everitt,
\newblock {\em Cluster Analysis},
\newblock John Wiley \& Sons, 2011.

\bibitem{hennig2015handbook}
Christian Hennig, Marina Meila, Fionn Murtagh, and Roberto Rocci,
\newblock {\em Handbook of Cluster Analysis},
\newblock Chapman and Hall/CRC, December 2015.

\bibitem{straub2015small}
Julian Straub, Trevor Campbell, Jonathan~P. How, and John~W. Fisher,
\newblock ``Small-variance nonparametric clustering on the hypersphere,''
\newblock in {\em 2015 IEEE Conference on Computer Vision and Pattern
  Recognition (CVPR)}, June 2015, pp. 334--342.

\bibitem{schroff2015facenet}
F.~Schroff, D.~Kalenichenko, and J.~Philbin,
\newblock ``Facenet: A unified embedding for face recognition and clustering,''
\newblock in {\em 2015 IEEE Conference on Computer Vision and Pattern
  Recognition (CVPR)}, June 2015, pp. 815--823.

\bibitem{wan2017ge2e}
Li~Wan, Quan Wang, Alan Papir, and Ignacio~Lopez Moreno,
\newblock ``Generalized end-to-end loss for speaker verification,''
\newblock {\em arXiv preprint arXiv:1710.10467}, 2017.

\bibitem{wang2017speaker}
Quan Wang, Carlton Downey, Li~Wan, Philip~Andrew Mansfield, and Ignacio~Lopez
  Moreno,
\newblock ``Speaker diarization with lstm,''
\newblock {\em arXiv preprint arXiv:1710.10468}, 2017.

\bibitem{buss2001spherical}
Samuel~R. Buss and Jay~P. Fillmore,
\newblock ``Spherical averages and applications to spherical splines and
  interpolation,''
\newblock {\em ACM Transactions on Graphics}, vol. 20, no. 2, pp. 95--126,
  2001.

\bibitem{kuhn1955hungarian}
Harold~W. Kuhn,
\newblock ``The hungarian method for the assignment problem,''
\newblock {\em Naval Research Logistics Quarterly}, vol. 2, pp. 83--97, 1955.

\end{thebibliography}

\end{document}